\documentclass[pdflatex,sn-apa]{sn-jnl}% APA Reference Style
\usepackage{graphicx}%
\usepackage{multirow}%
\usepackage{amsmath,amssymb,amsfonts}%
\usepackage{amsthm}%
\usepackage{mathrsfs}%
\usepackage[title]{appendix}%
\usepackage{xcolor}%
\usepackage{textcomp}%
\usepackage{manyfoot}%
\usepackage{booktabs}%
\usepackage{caption}
\usepackage{algorithm}%
\usepackage{algorithmicx}%
\usepackage{algpseudocode}%
\usepackage{listings}%
\usepackage{tabularx} 
\usepackage{graphicx} 
\usepackage{booktabs}
\usepackage{hyperref}
\usepackage{booktabs}
\usepackage{ltablex}

\usepackage{booktabs}
\usepackage{longtable}
\usepackage{array}

\usepackage{amsmath}
\usepackage{float}

\theoremstyle{thmstyleone}%
\theoremstyle{thmstyletwo}%

\theoremstyle{thmstylethree}%

\begin{document}

\title[Article Title]{Multi-Agent Learning Path Planning via LLMs}


% \author*[1,2]{\fnm{First} \sur{Author}}\email{iauthor@gmail.com}



% \affil*[1]{\orgdiv{Department}, \orgname{Organization}, \orgaddress{\street{Street}, \city{City}, \postcode{100190}, \state{State}, \country{Country}}}

% \affil[2]{\orgdiv{Department}, \orgname{Organization}, \orgaddress{\street{Street}, \city{City}, \postcode{10587}, \state{State}, \country{Country}}}

% \affil[3]{\orgdiv{Department}, \orgname{Organization}, \orgaddress{\street{Street}, \city{City}, \postcode{610101}, \state{State}, \country{Country}}}

% Authors
\author*[1]{\fnm{Haoxin} \sur{Xu}}\email{hxxu@shisu.edu.cn}

\author[2]{\fnm{Changyong} \sur{Qi}}\email{changyongqi@stu.ecnu.edu.cn}

\author[1]{\fnm{Tong} \sur{Liu}}\email{liutong@shisu.edu.cn}

\author[3]{\fnm{Bohao} \sur{Zhang}}\email{chambers4harr@163.com}

\author[2]{\fnm{Anna} \sur{He}}\email{51265901005@stu.ecnu.edu.cn}

\author[4]{\fnm{Longwei} \sur{Zheng}}\email{lwzheng@cityu.edu.mo}

\author[2]{\fnm{Xiaoqing} \sur{Gu}}\email{xqgu@ses.ecnu.edu.cn}

% Affiliations (only institutions)
\affil*[1]{Shanghai International Studies University}

\affil[2]{East China Normal University}

\affil[3]{Huaiyin Normal University}

\affil[4]{City University of Macau}

% % Authors
% \author*[1]{\fnm{Haoxin} \sur{Xu}}\email{hxxu@shisu.edu.cn}

% \author[2]{\fnm{Changyong} \sur{Qi}}\email{changyongqi@stu.ecnu.edu.cn}

% \author[1]{\fnm{Tong} \sur{Liu}}\email{liutong@shisu.edu.cn}

% \author[3]{\fnm{Bohao} \sur{Zhang}}\email{bhzhang@stu.ecnu.edu.cn}

% \author[3]{\fnm{Anna} \sur{He}}\email{51265901005@stu.ecnu.edu.cn}

% \author[4,5]{\fnm{Longwei} \sur{Zheng}}\email{lwzheng@cityu.edu.mo}

% \author[2]{\fnm{Xiaoqing} \sur{Gu}}\email{xqgu@ses.ecnu.edu.cn}

% % Affiliations
% \affil*[1]{\orgdiv{School of Education}, \orgname{Shanghai International Studies University}, \orgaddress{\street{1550 Wenxiang Road}, \city{Shanghai}, \postcode{201620}, \country{China}}}

% \affil[2]{\orgdiv{Lab of Artificial Intelligence for Education}, \orgname{East China Normal University}, \orgaddress{\street{3663 Zhongshan North Road}, \city{Shanghai}, \postcode{200062}, \country{China}}}

% \affil[3]{\orgdiv{School of Computer Science and Technology}, \orgname{East China Normal University}, \orgaddress{\street{3663 Zhongshan North Road}, \city{Shanghai}, \postcode{200062}, \country{China}}}

% \affil[4]{\orgdiv{School of Education}, \orgname{City University of Macau}, \orgaddress{\street{Avenida Padre Tomás Pereira}, \city{Macau}, \postcode{999078}, \country{China}}}

% \affil[5]{\orgdiv{State Key Laboratory of Cognitive Intelligence}, \orgname{}, \orgaddress{\city{Hefei}, \country{China}}}


\abstract{The integration of large language models (LLMs) into intelligent tutoring systems offers transformative potential for personalized learning in higher education. However, most existing learning path planning approaches lack transparency, adaptability, and learner-centered explainability. To address these challenges, this study proposes a novel Multi-Agent Learning Path Planning (MALPP) framework that leverages a role- and rule-based collaboration mechanism among intelligent agents, each powered by LLMs. The framework includes three task-specific agents: a learner analytics agent, a path planning agent, and a reflection agent. These agents collaborate via structured prompts and predefined rules to analyze learning profiles, generate tailored learning paths, and iteratively refine them with interpretable feedback. Grounded in Cognitive Load Theory and Zone of Proximal Development, the system ensures that recommended paths are cognitively aligned and pedagogically meaningful. Experiments conducted on the MOOCCubeX dataset using seven LLMs show that MALPP significantly outperforms baseline models in path quality, knowledge sequence consistency, and cognitive load alignment. Ablation studies further validate the effectiveness of the collaborative mechanism and theoretical constraints. This research contributes to the development of trustworthy, explainable AI in education and demonstrates a scalable approach to learner-centered adaptive instruction powered by LLMs.}

%%================================%%
%% Sample for structured abstract %%
%%================================%%




\maketitle

\section{Introduction}\label{sec1}

With the rapid development of online higher education, learners are increasingly confronted with cognitive overload and confusion when navigating complex knowledge systems. As a critical solution to these issues, Intelligent Tutoring Systems (ITS) have played an increasingly vital role in supporting personalized instruction. Within ITS, learning path planning is a key technique that aims to dynamically optimize learning trajectories and arrange the presentation sequence of learning resources based on learners’ individual needs, preferences, and prior knowledge~\citep{romero2019supporting,Amir2020survey}. A well-designed learning path can significantly improve learners' comprehension and long-term retention of knowledge, stimulate motivation and engagement, and ultimately contribute to better educational outcomes~\citep{Amir2020survey,Bucchiarone2023,SHI2020path}.

Despite its promising applications in intelligent education, learning path planning still faces several critical challenges in practice. First, it requires accurate identification of learners’ current knowledge states and preferences, along with continuous adaptation based on real-time feedback~\citep{ZHOU2018135,romero2019supporting}. However, traditional approaches often rely on predefined rules and lack adaptive flexibility, making them insufficient for addressing the diverse needs of individual learners~\citep{rahayu2023systematic,ZHOU2018135,ZHU2018102}. Second, mainstream approaches commonly lack explainability, making it difficult for learners and educators to understand the logic behind the recommended paths, thereby reducing trust in the system~\citep{app12083982,Diao2022Learning}. Third, in terms of evaluation, especially in offline scenarios, the absence of a unified metric framework makes it difficult to compare methods effectively, thus hindering further optimization and widespread deployment of path planning technologies~\citep{rahayu2023systematic,Amir2020survey,Dwivedi2018}.

In recent years, the rapid advancement of large language models (LLMs) and multi-agent systems has opened new possibilities for addressing these challenges~\citep{Wen2024Education,tran2025agentsurvey,li2024agent_survey,KASNECI2023edu_llm}. On the one hand, LLMs, with their powerful natural language understanding and generation capabilities, can perform diverse tasks under complex contexts and generate coherent, interpretable textual content, showing great potential in educational applications~\citep{Wen2024Education,Hu2024Plan}. On the other hand, multi-agent systems excel at task delegation, strategic coordination, and dynamic decision-making, making them particularly suitable for complex, information-dense, and hierarchical educational environments~\citep{tran2025agentsurvey,li2024agent_survey,gao2024large}.

Combining the strengths of these two technologies, LLM-based multi-agent systems can generate flexible learning paths according to learners’ goals, cognitive levels, and personal preferences, while providing interpretable explanations. However, accurately modeling learner-specific characteristics still requires large volumes of high-quality interaction data, and current LLMs face limitations in terms of efficiency and precision when conducting personalized modeling~\citep{wang2024survey_agent,guo2024largelanguagemodelbased}. To address this, auxiliary task-specific models such as academic risk prediction and knowledge tracing are needed to improve learner modeling accuracy. Furthermore, given the diversity and heterogeneity of learning resources, LLMs alone struggle to perform precise matching and optimal recommendations, limiting their ability to provide fine-grained personalized support. In this context, incorporating learning resource recommendation algorithms can help narrow the search space for LLMs and improve the accuracy of path generation~\citep{wang2024survey_agent,guo2024largelanguagemodelbased}.

To tackle the above challenges, this study proposes a multi-agent-based explainable learning path planning framework aimed at enabling more efficient and trustworthy personalized learning support. The method integrates key technologies such as academic risk identification, knowledge state perception, and learning resource recommendation, and employs division of labor among intelligent agents to dynamically generate learning paths. During path generation, we innovatively introduce Cognitive Load Theory (CLT) and the Zone of Proximal Development (ZPD) as pedagogical constraints, guiding agents to produce personalized paths that are both cognitively aligned and developmentally appropriate, along with detailed planning rationales. This method not only improves the interpretability and transparency of recommendations but also provides learners with actionable strategy suggestions, thereby enhancing their self-regulated learning ability.

This study makes the following key contributions:

\begin{itemize}
    \item We construct a complete technical framework covering academic risk detection, knowledge state modeling, resource recommendation, and learning path planning, enabling end-to-end personalized learning support and enhancing practical applicability.
    \item We propose an explainable learning path planning method based on multi-agent collaboration, generating intuitive and comprehensible explanations alongside the recommended paths to improve system transparency and user trust.
    \item We design a multi-dimensional evaluation framework grounded in CLT and ZPD, addressing the lack of standard metrics and supporting scientific assessment of path planning effectiveness.
    \item Experimental results on real-world learning datasets show that the proposed method can generate well-structured, logically coherent, and interpretable learning paths while effectively controlling learners’ cognitive load, resulting in improved learning efficiency and experience.
\end{itemize}

\section{Related Work}

\subsection{Learning Path Planning in Intelligent Tutoring Systems}

Learning path planning is a core capability in ITSs, designed to adaptively sequence learning resources according to learners’ knowledge levels, preferences, and pedagogical goals ~\citep{Amir2020survey,SHI2020path}. Early research primarily relied on rule-based strategies, such as decision trees and expert-authored knowledge maps, to generate fixed or semi-fixed learning paths ~\citep{CHI20097838}. These methods are interpretable and easy to implement but struggle to accommodate the dynamic nature of learners’ progress. To enhance adaptability, metaheuristic optimization approaches—including genetic algorithms, particle swarm optimization, and ant colony optimization—have been proposed ~\citep{ZHOU2018135}. While effective in producing diverse solutions, such methods often require extensive parameter tuning and lack pedagogical interpretability. More recently, deep learning and reinforcement learning models have been explored to model complex learner–content interactions and optimize long-term learning outcomes ~\citep{Bucchiarone2023}. However, these models typically function as black boxes, which reduces transparency and trust in educational settings. Despite significant advances, current learning path planning approaches still face challenges in real-time adaptation, scalability across subject domains, and the seamless integration of educational theories, motivating the exploration of hybrid solutions that combine adaptivity, scalability, and explainability.

\subsection{Explainability in Educational Recommender Systems}
Explainability has become a crucial requirement in Educational Recommender Systems (ERS) to promote transparency, trust, and learner engagement ~\citep{MILLER20191,10.1145/3490099.3511140}. Early explainable ERS approaches leveraged rule-based logic or content metadata to present clear rationale for recommendations, such as prerequisite relationships between resources ~\citep{10.1145/506218.506247}. More recent developments have focused on knowledge graph-based models that explicitly encode semantic relationships between concepts, enabling path explanations grounded in structured knowledge ~\citep{Wang_Wang_Xu_He_Cao_Chua_2019}. Visualization-based methods have also emerged, offering learners graphical overviews of their progress and the reasoning behind recommendations ~\citep{10.1145/3314183.3323463}. Additionally, post-hoc model-agnostic techniques, such as SHAP or LIME, have been adapted to educational contexts to interpret black-box models ~\citep{INR-066}. Despite these advances, most ERS explanations remain detached from the underlying pedagogical theories, limiting their relevance for learning design. Integrating cognitive principles such as Cognitive Load Theory (CLT) and the Zone of Proximal Development (ZPD) into explanation generation could provide learners and educators with not only transparency but also actionable, theory-grounded guidance for effective learning interventions.

\subsection{Large Language Models in Education}
LLMs such as GPT-4, LLaMA, and Claude have demonstrated exceptional capabilities in natural language understanding, reasoning, and generation, prompting significant interest in their educational applications ~\citep{KASNECI2023edu_llm,Adeshola25112024}. LLMs have been used to build intelligent tutoring systems capable of adaptive question answering, scaffolded explanations, and contextual feedback ~\citep{Wen2024Education}. They have also been applied in automated grading ~\citep{10.1145/3706468.3706481}, personalized content generation ~\citep{chen2024large}, and educational dialogue systems. Their versatility enables them to support diverse learning scenarios, from language learning to STEM problem solving. However, LLM deployment in education faces challenges in aligning generated content with curriculum standards, maintaining factual accuracy, and adapting recommendations to learners’ cognitive profiles. Furthermore, most current implementations rely on static prompts, limiting adaptability to evolving learner states. Integrating LLMs with structured pedagogical frameworks such as CLT and ZPD, and embedding them in multi-agent workflows, could improve personalization, ensure theoretical grounding, and enhance both transparency and learning effectiveness.

\subsection{Multi-Agent Systems in Education}
Multi-agent systems (MAS) enable distributed problem solving through the collaboration of autonomous agents, each specializing in specific tasks such as learner modeling, resource recommendation, or assessment ~\citep{guo2024largelanguagemodelbased,article}. In educational contexts, MAS have been applied to adaptive tutoring, collaborative learning, and simulation-based training. Early MAS-based tutoring systems assigned fixed roles to agents, such as monitoring progress or delivering feedback, often using pre-defined rules for coordination. More recent approaches leverage reinforcement learning and communication protocols to allow agents to dynamically adapt their strategies and share knowledge in real time ~\citep{LUO2025100852}. MAS can also facilitate integration with external data sources, enabling cross-domain personalization and scalability across diverse curricula. However, existing MAS implementations in education often lack deep integration with pedagogical theories, limiting their ability to optimize learning pathways for cognitive and motivational outcomes. Combining MAS with large language models offers an opportunity to enhance reasoning, negotiation, and explainability, while grounding agent decision-making in frameworks such as CLT and ZPD.

Although significant progress has been made in learning path planning, explainable educational recommenders, large language model integration, and multi-agent systems, several research gaps remain. First, most learning path planning approaches either focus on adaptivity without explainability or provide explanations that are detached from pedagogical principles such as CLT and the ZPD ~\citep{MILLER20191}. Second, while LLMs offer strong natural language reasoning and generation capabilities, their use in dynamic, multi-stage educational workflows remains limited ~\citep{KASNECI2023edu_llm,Adeshola25112024}. Third, existing MAS-based educational systems rarely combine LLM reasoning with structured educational theory, resulting in suboptimal alignment between system recommendations and learners’ cognitive states ~\citep{guo2024largelanguagemodelbased}. To address these gaps, this work proposes a Multi-Agent Learning Path Planning (MALPP) framework that integrates LLM-driven reasoning with MAS collaboration and theory-grounded constraints. By embedding CLT and ZPD into the path generation and reflection process, MALPP enhances transparency, adaptivity, and cognitive alignment, offering a comprehensive solution for personalized, explainable learning support.

\section{Methodology}

\subsection{Problem Formulation and Evaluation Metrics}
%2.2.1 问题建模
In intelligent tutoring systems, learning path planning aims to dynamically generate personalized learning trajectories based on learners' individual needs, preferences, and prior knowledge, thereby enhancing learning efficiency and experience. To ensure rigor and consistency throughout the research, this section defines key terms involved in learning path planning and establishes a unified semantic framework for the study. We formulate the learning path planning task as a constrained optimization problem. The inputs include learner profile information, learning resources, and a knowledge graph. The objective is to identify an optimal learning path that maximizes the learner's overall learning effectiveness while satisfying specific constraints.

Specifically, the learner \( u_i \) is represented by a feature vector \( S_i \in \mathbb{R}^d \), where \( d \) denotes the dimensionality of the vector. The vector encodes personalized attributes such as demographic features, knowledge mastery levels, and learning preferences. The set of available learning resources is denoted as \( C = \{c_1, c_2, \dots, c_n\} \), where each element \( c_i \) represents a learning unit, such as a video, document, or practice exercise. For each resource \( c_i \), we define the expected learning duration as \( t_i \), indicating the time required to complete the resource, and the estimated learning effectiveness as \( E_i \), which reflects the potential knowledge gain or skill improvement the learner may obtain from engaging with the resource. Domain knowledge is modeled using a knowledge graph \( G \), which captures both the ontological attributes of knowledge points (e.g., names and types) and the prerequisite relationships among them (e.g., the topological structure of prerequisite and successor concepts).

Based on the elements above, the goal of learning path planning is to generate an ordered sequence of learning units \( P = (p_1, p_2, \dots, p_k) \), given the learner profile \( S_i \), the resource set \( C \), and the knowledge graph \( G \). The path should maximize the cumulative learning effectiveness under constraints such as time limits and knowledge dependencies. The formal objective is defined in Equation~\eqref{path_obj}, where \( P^* \) denotes the optimal learning path, \( k \) is the number of units in the path, \( p_j \) is the \( j \)-th learning unit, \( E_{p_j} \) is its associated learning effectiveness, and \( \mathcal{F}(\cdot) \) is the path generation function that produces \( P \) from the given inputs \( S_i \), \( C \), and \( G \). The optimization seeks the path with the highest cumulative effectiveness from the feasible path set:

\begin{equation}\label{path_obj}
    P^* = \arg\max_{P = (p_1, p_2, \dots, p_k)} \sum_{j=1}^{k} E_{p_j}, \quad \text{s.t. } P = \mathcal{F}(S_i, C, G)
\end{equation}

\subsection{Overall Framework Overview}
Traditional learning path planning methods often suffer from limited explainability, poor adaptability, and lack of robust offline evaluation mechanisms, making them inadequate for meeting the demands of personalized learning. To address these limitations, we propose a novel framework called Multi-Agent Learning Path Planning (MALPP), which aims to enhance the transparency, adaptability, and dynamic optimization capabilities of learning path generation.

\begin{figure}[h]
  \centering
  \includegraphics[width=\columnwidth]{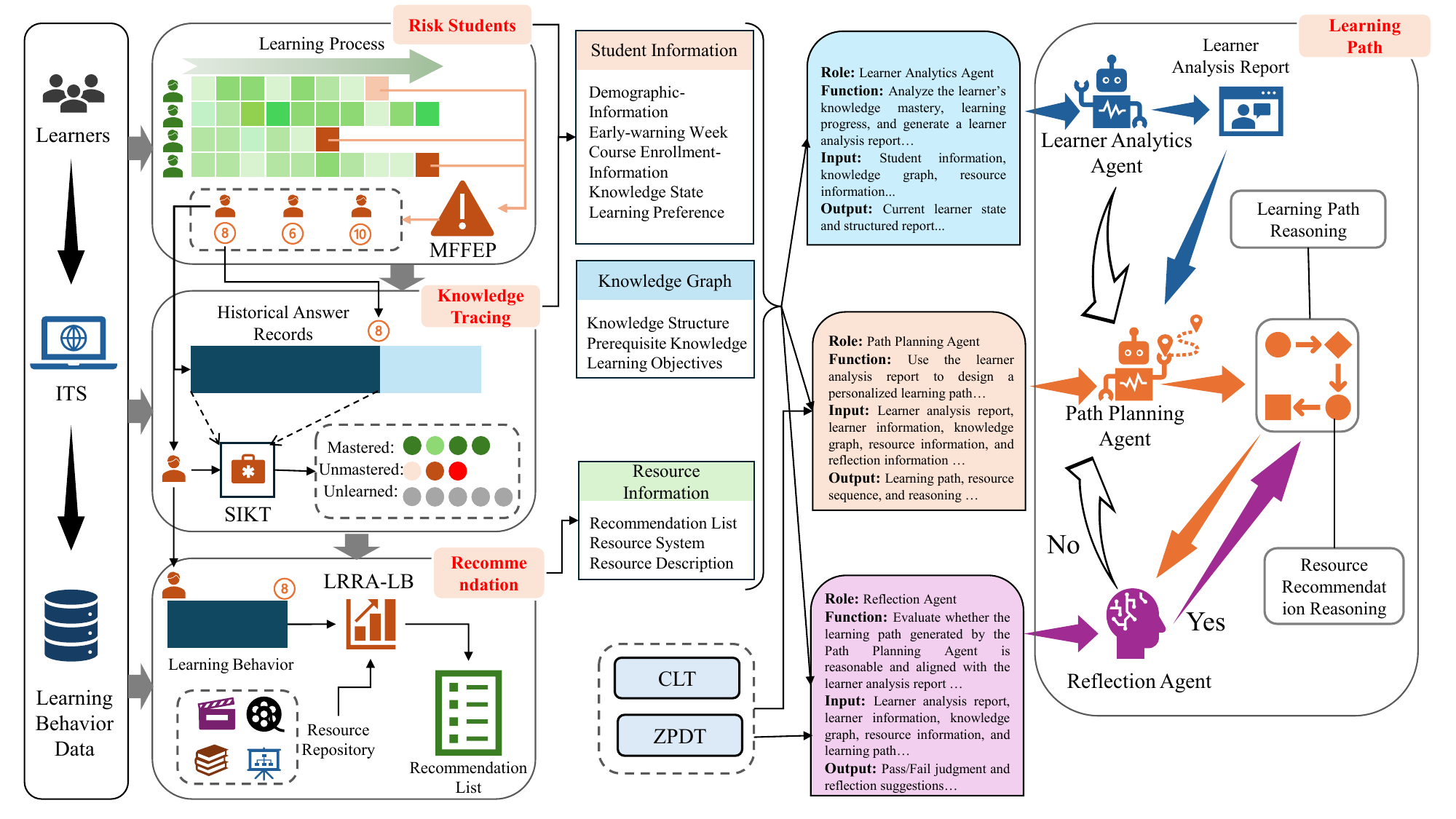}
  \caption{Multi-agent-based explainable learning path planning framework}
  \label{learn_path_llm}
\end{figure}

As illustrated in Figure~\ref{learn_path_llm}, the proposed approach integrates multiple core components of intelligent tutoring systems into a cohesive pipeline. The process begins with an academic risk prediction model, which identifies learners who may be at risk of academic underperformance. For these at-risk learners, a knowledge tracing model is applied to assess and continuously track their current knowledge states. Based on this diagnostic information, a learning resource recommendation algorithm generates a preliminary list of candidate learning resources.

Building upon this foundation, a set of intelligent agents collaborate to plan a personalized learning path. First, the learner analytics agent analyzes the learner's profile and generates a detailed diagnostic report. Next, the path planning agent utilizes this report to produce an initial learning path along with a set of justifications for its decisions. The proposed path is then evaluated by the reflection agent, which conducts a rigorous assessment. If the path does not meet evaluation criteria, the reflection agent provides revision suggestions. The planning agent incorporates this feedback and iteratively refines the path until it satisfies the reflective evaluation.

By integrating intelligent agents with task-specific models and orchestrating them within a feedback-driven loop, the proposed method enables a data-driven and interpretable learning path planning process. It ensures that the generated paths are pedagogically adaptive, transparent in rationale, and optimized through dynamic iteration. The following sections detail each component of the framework.

\subsubsection{Learning Path Planning Information Extraction}

To enable accurate and adaptive learning path planning, it is essential to integrate multiple sources of data and extract key information relevant to the planning process. This procedure includes the following four steps:

(1) Academic risk identification.  
During the course process, the academic risk prediction model MFFEP is employed to continuously monitor students. Once a student \( u_i \) is identified as being at academic risk, the system immediately generates a risk alert and records the alert time \( t_i \).

(2) Knowledge state diagnosis.  
After detecting the at-risk student \( u_i \) and the corresponding alert time \( t_i \), the knowledge tracing model SIKT is used to assess the student's knowledge state \( K_i = \{k_{i,1}, k_{i,2}, \dots, k_{i,m}\} \), where \( k_{i,j} \) represents the mastery level of student \( u_i \) on knowledge point \( j \) at time \( t_i \), and \( m \) is the total number of knowledge points. By dynamically updating \( K_i \), the system can identify knowledge deficiencies and generate a basis for personalized learning path construction.

(3) Personalized learning resource recommendation.  
Based on the diagnosed knowledge state, the learning resource recommendation algorithm LRRA-LBP is applied to retrieve learning materials that best match the student's current needs from the existing resource pool. A personalized list of recommended learning resources is generated as \( R_i = \{r_{i,1}, r_{i,2}, \dots, r_{i,n}\} \), where \( r_{i,j} \) denotes the \( j \)-th resource recommended for student \( u_i \), and \( n \) is the number of recommended resources.

(4) Extraction of supplementary information for path planning.  
In addition to the information obtained above, additional key data must be extracted from the intelligent tutoring system to support the learning path planning process. This includes learner profile information \( S_i \), which contains attributes such as age, gender, grade level, and major, depending on the structure of the system database. It also includes the knowledge graph \( G \), which encodes the names, structures, and learning objectives of knowledge points, enabling the system to respect prerequisite relationships and generate pedagogically coherent learning sequences. Finally, metadata associated with the recommended resources \( R_i \) is also required, including resource titles, descriptions, and their corresponding knowledge points, to ensure accurate matching between learner needs and learning content.

The extracted information above forms the foundation for the learning path planning task and supports the collaborative execution of subtasks by different agents in the multi-agent system.

\subsubsection{Multi-Agent Collaboration Mechanism}

\begin{figure}[h]
  \centering
  \includegraphics[width=0.9\columnwidth]{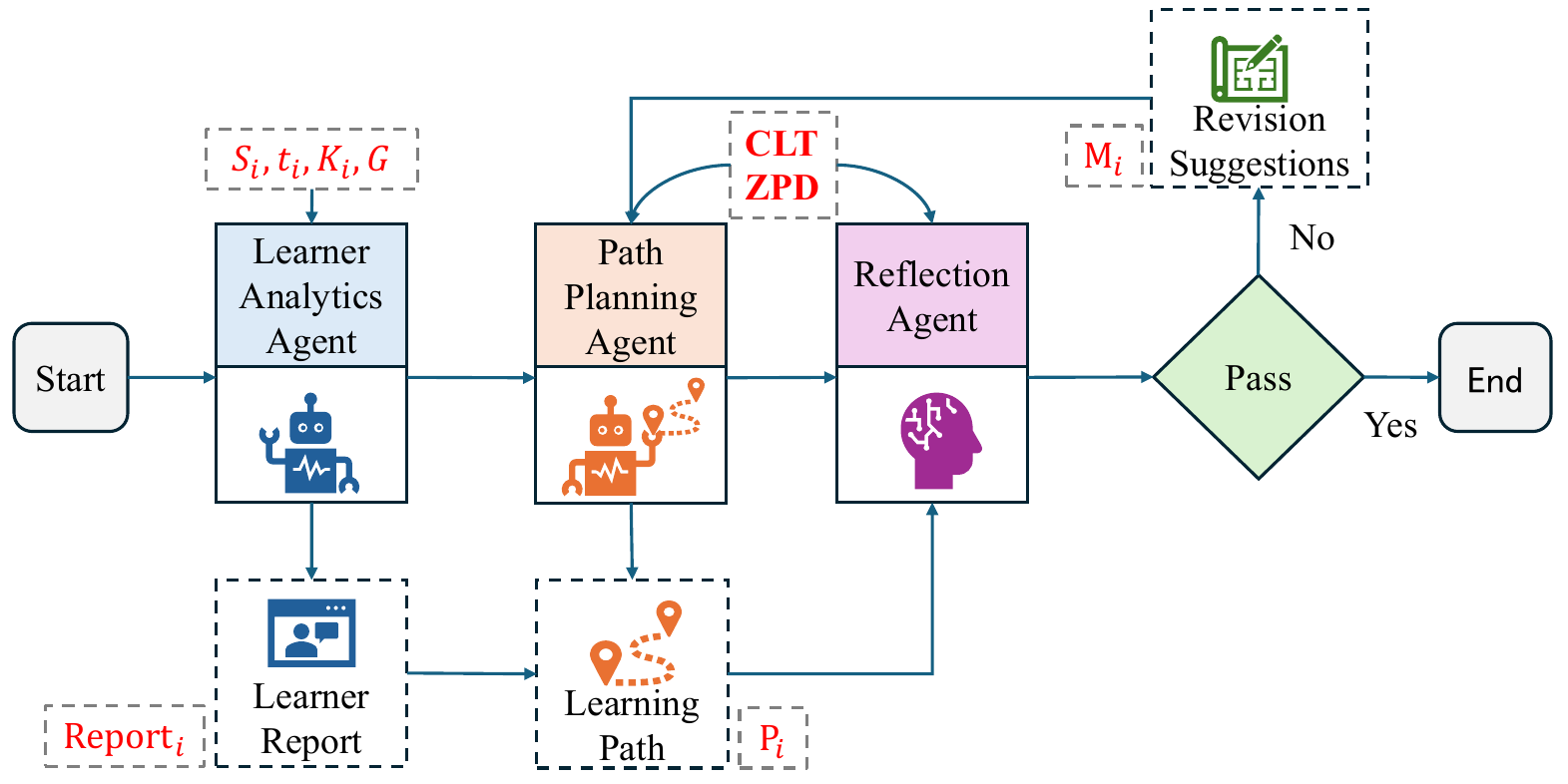}
  \caption{Multi-agent collaboration mechanism for learning path planning}
  \label{learn_path_work}
\end{figure}

To ensure that the generated learning path is both pedagogically reasonable and cognitively adaptive, we propose a multi-agent collaboration mechanism, as illustrated in Figure~\ref{learn_path_work}. In this mechanism, multiple agents work in coordination to complete the learning path planning task. Given the domain-specific nature of path planning, the collaboration strategy adopts a role- and rule-based design: each agent operates according to its predefined role while adhering to a set of predefined rules for information exchange and task delegation. This enables efficient division of labor and collaborative execution. The process proceeds as follows.

The collaboration process among agents proceeds as follows. The learner analytics agent first receives the extracted information from the previous stage, including \( S_i, t_i, K_i, G \), and analyzes the learner's current state. Based on this analysis, it generates a personalized diagnostic report \( Report_i \). The path planning agent then takes \( Report_i \) as input and produces an initial learning path \( P_i \). This path is subsequently evaluated by the reflection agent. If the path does not meet certain evaluation criteria, the agent generates revision suggestions \( M_i \); otherwise, the path is accepted as the final output. If suggestions \( M_i \) are provided, the planning agent incorporates the feedback and adjusts the path accordingly, generating a revised version \( P_{i+1} \).

To prevent infinite loops during the iterative generation and revision process, the system imposes a maximum of three rounds for the planning-reflection-revision cycle. If the path still fails to pass the reflection evaluation after three iterations, the final version \( P_{i+2} \) is adopted by default. In the following sections, we describe the detailed functions and implementation mechanisms of each agent.

\subsubsection{Learner Analytics Agent}

\begin{figure}[h]
  \centering
  \includegraphics[width=\columnwidth]{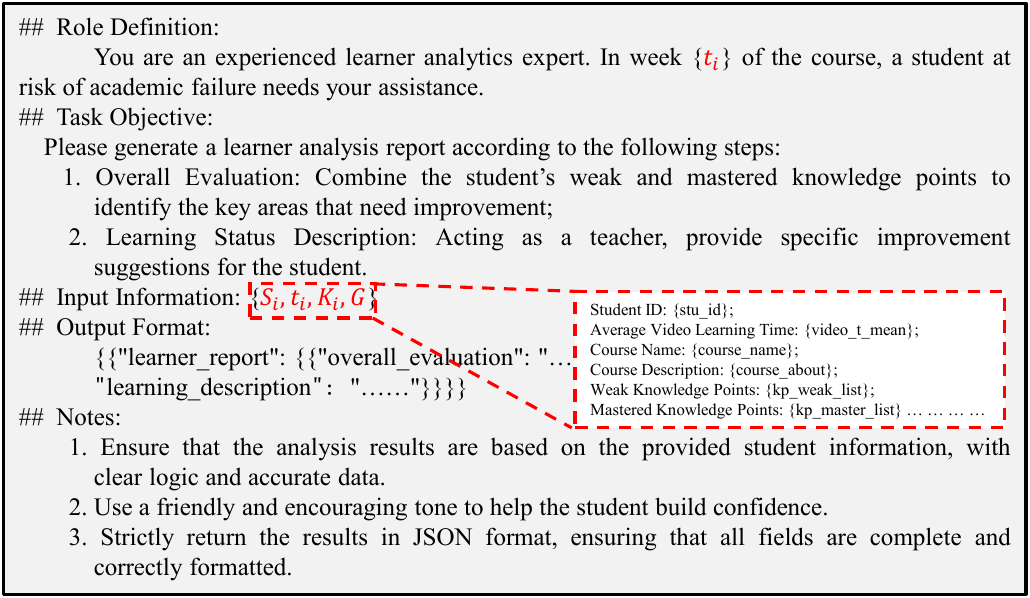}
  \caption{Prompt design of the learner analytics agent}
  \label{learn_path_ana}
\end{figure}

After collecting the necessary information, the learner analytics agent is responsible for accurately analyzing and summarizing the learner’s current state in order to support subsequent path planning. Specifically, this agent constructs a dynamic prompt template \( Prompt_a \) that integrates the available input data. The structure of this prompt is illustrated in Figure~\ref{learn_path_ana}.

The role of the learner analytics agent is assigned to that of an experienced educational diagnostician to ensure the accuracy and relevance of the analysis. The dynamic inputs to the prompt (indicated in red in the figure) include the identified at-risk student \( U_i \), the corresponding profile information \( \{S_i, t_i, K_i\} \), and the domain knowledge structure \( G \). The agent processes this information to generate a comprehensive diagnostic report on the learner’s current academic status. The final output is a JSON-formatted learner report \( Report_i \), which contains information such as knowledge mastery, potential weaknesses, and personalized learning preferences. This serves as a reliable foundation for subsequent learning path generation.

\subsubsection{Path Planning Agent}

\begin{figure}[h]
  \centering
  \includegraphics[width=\columnwidth]{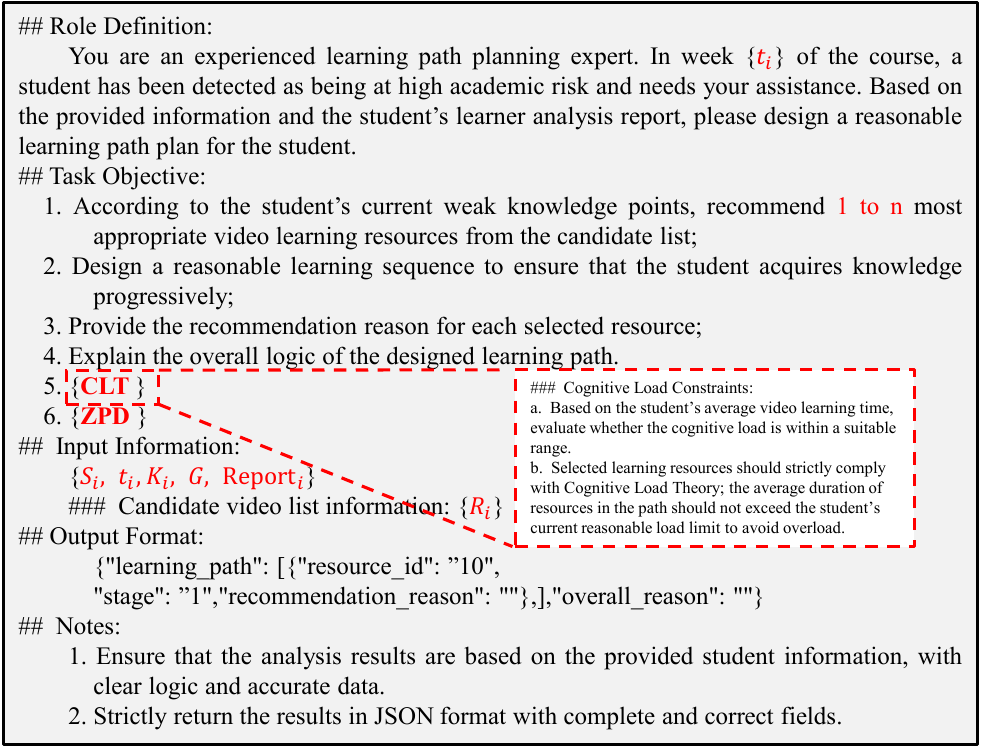}
  \caption{Prompt design of the path planning agent}
  \label{learn_path_plan}
\end{figure}

After receiving the learner report \( Report_i \), the path planning agent is responsible for generating a feasible and pedagogically appropriate learning path for the target student. As shown in Figure~\ref{learn_path_plan}, the agent constructs a dynamic prompt template \( Prompt_p \) to guide the generation of a personalized learning plan.

This agent is assigned the role of an experienced expert in learning path design, with the main objective of assisting student \( u_i \), who was identified as being at academic risk in week \( t_i \), by generating a reasonable learning path along with detailed justifications. During the planning process, the agent is guided by two key constraint modules. The first is the Cognitive Load Theory (CLT) constraint module (see Figure~\ref{learn_path_plan}), which ensures that the learning path does not exceed the learner’s cognitive capacity, thereby preventing overload and promoting retention. The second is the Zone of Proximal Development (ZPD) constraint module (see Figure~\ref{learn_path_ref}), which guarantees that the order of knowledge acquisition follows a progressive trajectory, expanding on what the learner already knows to improve learning effectiveness.

The input to the agent includes the learner profile and contextual data \( \{S_i, t_i, K_i\} \), the knowledge graph \( G \), the learner report \( Report_i \), and the recommended resource list \( R_i \). Together, these inputs provide the necessary information for generating a personalized path. The output is a JSON-formatted structure containing the planned learning path \( P_i \) and a global rationale. Each node in the path corresponds to a specific learning resource and includes a localized explanation to enhance interpretability.

\subsubsection{Reflection Agent}

\begin{figure}[h]
  \centering
  \includegraphics[width=\columnwidth]{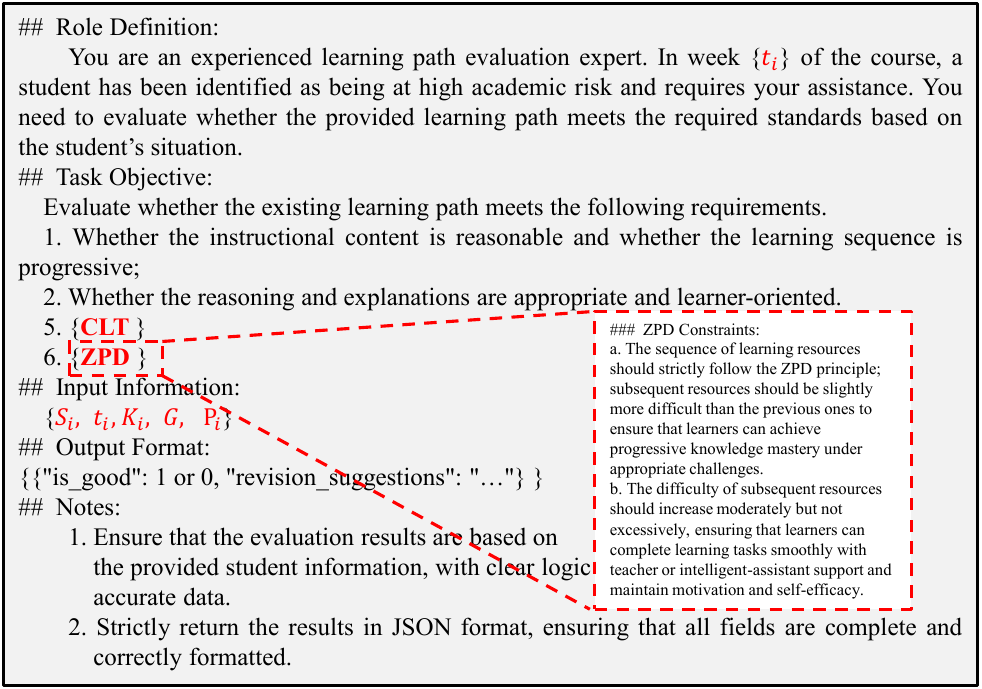}
  \caption{Prompt design of the reflection agent}
  \label{learn_path_ref}
\end{figure}

After the learning path \( P_i \) is generated, it is evaluated by the reflection agent, whose task is to assess the pedagogical appropriateness of the proposed path. As shown in Figure~\ref{learn_path_ref}, this agent uses a dynamic prompt template \( Prompt_r \) to carry out the evaluation process.

The reflection agent is assigned the role of an experienced expert in learning path evaluation. Its primary responsibility is to determine whether the current learning path adheres to key educational principles such as cognitive load control and progressive knowledge development. To support this, the agent incorporates two constraint modules: the Cognitive Load Theory (CLT) module and the Zone of Proximal Development (ZPD) module. The CLT module evaluates whether the total cognitive load of the path is within the learner’s acceptable range, helping to avoid the negative effects of overload. The ZPD module ensures that the sequence of learning steps follows a gradual progression aligned with the learner’s current level of understanding.

The inputs to the reflection agent include the learner's contextual data \( \{S_i, t_i, K_i\} \), the knowledge graph \( G \), and the proposed learning path \( P_i \). The output is a JSON-formatted structure. If the path passes the evaluation, it is accepted as the final version. If it fails, a set of revision suggestions \( M_i \) is generated and returned to the path planning agent for further refinement.

\section{Experiment Settings}

\subsection{Dataset}
To conduct a comprehensive study on learning path planning, it is essential to select an appropriate dataset. After a thorough analysis of several publicly available educational datasets, we found that many of them exhibit common limitations. For instance, some datasets only represent knowledge points using IDs without meaningful names, or they focus solely on learners' answering records while lacking essential behavioral data. After comparison, we selected the MOOCCubeX dataset~\citep{yu2021mooccubex} for our experiments.

MOOCCubeX is a widely used large-scale MOOC dataset\footnote{Dataset source: \url{https://github.com/THU-KEG/MOOCCubeX}} containing data on 3.33 million learners and 3,240 courses. It provides course titles, chapter-level content, and partial knowledge structure information that can support reasoning tasks performed by large language models. Moreover, it includes diverse types of learner interaction data, such as video views, exercise responses, and forum comments, which are essential for supporting the tasks of academic risk prediction, knowledge diagnosis, and resource recommendation in our proposed framework.

However, the dataset also presents certain limitations, including a limited variety of learning behaviors, the absence of demographic information about learners, and an imbalanced distribution of answer records.

% Due to the token cost involved in operating our multi-agent LLM-based framework and limited computational resources, we could not evaluate the entire dataset. 
Therefore, we constructed a representative subset from MOOCCubeX for detailed analysis. Specifically, we selected the top 20 courses based on video viewing frequency and the top 20 courses based on answering frequency. Taking the intersection of both sets resulted in 6 representative courses. After preprocessing and data cleaning, the final dataset includes 1,453 learners, 428 instructional videos, 209 exercises, 446 knowledge points, 77,498 video-view records, 7,489 comment records, 1,350 questions, and 197,085 answer logs. Table~\ref{course_info} provides a summary of the six selected courses.

For effective learning path planning, the intelligent agents require explicit knowledge point information. However, such information is not directly available in MOOCCubeX. To address this, we extracted knowledge point names from chapter descriptions using information extraction techniques, followed by manual verification. Based on this, we constructed knowledge graphs for all selected courses. 
For example, the course "Fundamentals of Mechanical Design" is organized into 15 chapters, each containing multiple knowledge points, which serve as the structural foundation for subsequent learning path generation.
% An example knowledge graph for the course "Fundamentals of Mechanical Design" is shown in Table~\ref{knowledge_structure}. The course is 

\begin{table}[h]
\caption{Course statistics in the MOOCCubeX dataset}
\centering
\small
\begin{tabularx}{\textwidth}{lXXXX}
\toprule
\textbf{Course Title} & \textbf{Learners} & \textbf{Video Records} & \textbf{Answer Records} & \textbf{Resources} \\
\midrule
Fundamentals of Mechanical Design & 101  & 17,578  & 13,373 & 85   \\
Logistics Task Organization and Execution & 98   & 7,533   & 23,343 & 119  \\
Fundamentals of Analog Electronics         & 198  & 5,318   & 48,319 & 213  \\
Academic English for STEM Majors           & 224  & 10,418  & 34,489 & 105  \\
Spirit of the War to Resist US Aggression  & 609  & 25,565  & 54,214 & 60   \\
Engineering Ethics and Academic Integrity  & 223  & 11,086  & 23,347 & 54   \\
\bottomrule
\end{tabularx}
\label{course_info}
\end{table}

\subsection{Baseline Models}

The performance of intelligent agents in our framework depends heavily on the capabilities of the underlying LLMs. To identify the most suitable foundation models for our learning path planning task, we selected and evaluated seven widely used LLMs: GPT-3.5, GPT-4o, Llama3.3, Claude-3, DeepSeek-V3, Qwen-Plus, and XXNU-Plus. These models vary in origin (open-source vs. closed-source; international vs. domestic) and scale, ranging from 70B to 671B parameters. Detailed information is presented in Table~\ref{all_llms}.

Based on these foundation models, we implemented and compared three learning path planning approaches:

\begin{itemize}
\item RBM (Random Baseline Model): This model randomly selects \( N \) resources from the recommended resource list as the learning path. It does not leverage any learner features or content understanding, and it is unable to provide explanations for the generated paths. It serves as a naive random baseline.

\item SLMLPP (Single Large Model for Learning Path Planning): This model does not use a multi-agent structure. Instead, it directly constructs a prompt using the extracted information from the first three steps and queries a single LLM to generate the learning path and explanations. Unlike our proposed method, SLMLPP lacks modularity and the task decomposition benefits of agent-based collaboration.

\item MALPP: This is the proposed Multi-Agent Learning Path Planning method. It utilizes multiple agents to collaboratively perform learner modeling, resource matching, path optimization, and explanation generation, thereby improving the accuracy and adaptability of the resulting learning paths.
\end{itemize}

\begin{table}[t] % 用 t/b/h，不用 H
\caption{Foundation Models for Learning Path Planning}
\centering
\small
\begin{tabularx}{\textwidth}{
  >{\raggedright\arraybackslash}p{2.6cm}
  >{\raggedright\arraybackslash}X
  >{\centering\arraybackslash}p{2.4cm}
  >{\centering\arraybackslash}p{2.2cm}}
\toprule
\textbf{Model Name} & \textbf{Description} & \textbf{Parameters} & \textbf{Release Date} \\
\midrule
\href{https://openai.com/research/gpt-3}{GPT-3.5} &
An improved version of GPT-3 developed by OpenAI, with enhanced language understanding and generation capabilities. &
175B & 2022.11 \\
\href{https://openai.com/research/gpt-4}{GPT-4o} &
The omnipotent version of GPT-4 by OpenAI, featuring enhanced memory and conversational abilities. &
$\geq$\,175B & 2024.05 \\
\href{https://ai.facebook.com/research/llama}{Llama3.3} &
The third-generation model in Meta’s open-source Llama series, aimed at improving language understanding and generation. &
70B & 2024.12 \\
\href{https://www.anthropic.com/claude}{Claude-3} &
A large model by Anthropic, designed to create safe and reliable AI systems aligned with human values. &
$>100$B & 2024.12 \\
\href{https://qwen.alibaba.com}{Qwen-Plus} &
An AI model developed by Alibaba, focused on enhancing mathematical reasoning, coding, and problem-solving skills. &
$>100$B & 2024.10 \\
\href{https://deepseek.com/models/v3}{DeepSeek-V3} &
An open-source large language model by DeepSeek, employing a mixture-of-experts architecture for efficient reasoning and generation. &
671B & 2024.12 \\
\href{https://ai.xxnu.edu.cn/models/plus}{XXNU-Plus} &
An intelligent chatbot developed by XXNU, integrating capabilities from SparkDesk, QWen, ChatGLM, DeepSeek, and more. &
-- & 2024.12 \\
\bottomrule
\end{tabularx}
\label{all_llms}
\end{table}

\subsection{Evaluation Metrics}

To comprehensively evaluate the effectiveness of learning path planning algorithms, we propose four key evaluation metrics based on previous studies~\citep{Hossein2020path}. These metrics include Average Path Length (APL), Average Learning Duration (ALD), Cognitive Load Misalignment Rate (CLMR), and Knowledge Sequence Consistency (KSC). APL and ALD are designed to describe the stylistic tendencies of different methods in generating learning paths. APL reflects the overall length of the path, while ALD measures the average estimated learning time per resource in the path. These indicators help compare whether a method tends to recommend shorter or longer paths and lighter or heavier learning loads. CLMR is used to assess whether the generated learning path aligns with the learner’s cognitive load capacity. A lower CLMR indicates that the path is better calibrated to avoid cognitive overload or underload while maintaining learning effectiveness. KSC evaluates whether the order of learning resources in the path respects the logical dependencies among knowledge points. Higher KSC values indicate better adherence to pedagogical sequencing, which is essential for ensuring the path’s instructional coherence and learning efficiency. Together, these four metrics allow for a well-rounded comparison of different approaches in terms of both performance and planning style, ensuring the scientific rigor and credibility of the evaluation process.

\subsubsection{Average Path Length (APL)}

APL measures the average number of learning resources (e.g., videos, texts, exercises) included in a learning path. It reflects the complexity and coverage of the generated paths. Let \( N \) be the total number of learning paths, and let \( |P_i| \) denote the length in path \( P_i \). The APL is calculated as follows:

\begin{equation}\label{APL}
    APL = \frac{\sum_{i=1}^{N} |P_i|}{N}
\end{equation}

\subsubsection{Average Learning Duration (ALD)}

ALD measures the average estimated time learners are expected to spend on each resource within a learning path. It reflects the average time investment required per learning item. Let \( |P_i| \) be the number of learning resources in path \( P_i \), and let \( T_j \) represent the estimated learning duration of the \( j \)-th resource. The average learning duration for a single path \( P_i \), denoted as \( ALD_i \), is calculated as:

\begin{equation}\label{ALD_i}
    ALD_i = \frac{\sum_{j=1}^{|P_i|} T_j}{|P_i|}
\end{equation}

The overall ALD across all \( N \) learning paths is computed as the average of all \( ALD_i \) values:

\begin{equation}\label{ALD}
    ALD = \frac{\sum_{i=1}^{N} ALD_i}{N}
\end{equation}

\subsubsection{ Cognitive Load Misalignment Rate (CLMR)}

According to Cognitive Load Theory, human working memory has limited capacity and can only process a finite amount of information at one time. When the cognitive demands exceed this limit, learning efficiency declines significantly. Therefore, it is important to consider cognitive constraints during learning path planning to ensure that the recommended tasks do not exceed the learner’s processing ability. To this end, we propose the Cognitive Load Misalignment Rate (CLMR) to evaluate whether the generated learning path matches the learner’s cognitive load level. The computation proceeds as follows.

The total cognitive load of a learner \( u_i \), denoted as \( CL_{\text{student}}^i \), consists of three components: intrinsic cognitive load \( CL_{\text{int}}^i \), germane cognitive load \( CL_{\text{ger}}^i \), and extraneous cognitive load \( CL_{\text{ext}} \). The intrinsic load is calculated as the learner’s average time spent per learning resource (Equation~\eqref{cl_int}), where \( T_{ij} \) is the time that learner \( u_i \) spent on resource \( j \), and \( m \) is the number of resources they have studied.

\begin{equation}\label{cl_int}
CL_{\text{int}}^i = \frac{1}{m} \sum_{j=1}^{m} T_{ij}
\end{equation}

The germane load reflects meaningful engagement in knowledge construction and is estimated using the average daily time the learner spends participating in discussion forums (Equation~\eqref{cl_ger}), where \( d \) is the total number of learning days and \( F_{i,h} \) is the time learner \( i \) spent in forums on day \( h \).

\begin{equation}\label{cl_ger}
CL_{\text{ger}}^i = \frac{1}{d} \sum_{h=1}^{d} F_{i,h}
\end{equation}

The extraneous cognitive load \( CL_{\text{ext}} \) arises from system design or interface factors. In a consistent learning system environment, it can be treated as a constant. Thus, the learner’s total cognitive load is calculated as:

\begin{equation}\label{stu_cl}
CL_{\text{student}}^i = CL_{\text{int}}^i + CL_{\text{ger}}^i + CL_{\text{ext}}
\end{equation}

For a given learning path \( P_i \), its cognitive load \( CL_{\text{path}}^i \) is estimated using the average historical time required for each resource in the path (Equation~\eqref{cl_path}). The average time per resource \( CL_{\text{res}}^j \) is defined in Equation~\eqref{cl_res}, where \( n \) is the number of learners who have studied resource \( Res_j \), and \( T_{i,j} \) is the time spent by learner \( u_i \) on that resource.

\begin{equation}\label{cl_path}
CL_{\text{path}}^i = \frac{1}{|P_i|} \sum_{j=1}^{|P_i|} CL_{\text{res}}^j
\end{equation}
\begin{equation}\label{cl_res}
CL_{\text{res}}^j = \frac{1}{n} \sum_{i=1}^{n} T_{i,j}
\end{equation}

For each learner \( u_i \) and learning path \( P_i \), we define the misalignment rate \( CLMR^i \) as shown in Equation~\eqref{clmr_i}. If \( CLMR^i > 0 \), the path load is below the learner's cognitive capacity, indicating that the path may be too simple; conversely, if \( CLMR^i < 0 \), the path may be too difficult and thus unmanageable for the learner.

\begin{equation}\label{clmr_i}
CLMR^i = \frac{CL_{\text{student}}^i - CL_{\text{path}}^i}{CL_{\text{student}}^i}
\end{equation}

Finally, we compute the overall CLMR across all \( N \) learners by averaging the absolute misalignment values, as shown in Equation~\eqref{clmr}. The resulting value \( CLMR \in [0,1] \), with lower values indicating better alignment between learning path complexity and learner cognitive capacity.

\begin{equation}\label{clmr}
CLMR = \frac{1}{N} \sum_{i=1}^{N} |CLMR^i| = \frac{1}{N} \sum_{i=1}^{N} \frac{|CL_{\text{student}}^i - CL_{\text{path}}^i|}{CL_{\text{student}}^i}
\end{equation}

\subsubsection{Knowledge Sequence Consistency (KSC)}

According to the theory of the Zone of Proximal Development (ZPD), the difficulty of knowledge points in a learning path should gradually increase to align with learners’ evolving cognitive readiness. We define KSC (Knowledge Sequence Consistency) as a metric to assess whether the sequence of knowledge points in a learning path exhibits a strictly increasing difficulty pattern, thereby reflecting a progressive learning logic. The consistency score for a single learning path \( P_j \), denoted as \( KSC_j \), is computed as follows (Equation~\eqref{ksc}). If the sequence of knowledge points in the path satisfies \( K_{j,i} < K_{j,i+1} \), where \( K_{j,i} \) represents the difficulty or hierarchical level of the \( i \)-th knowledge point, a score of 1 is awarded; otherwise, a score of 0 is assigned. If a path contains only one knowledge point, we define \( KSC_j = 0.5 \).

\begin{equation}\label{ksc}
KSC_j = \begin{cases} 
\frac{1}{|P_j|} \sum_{i=1}^{|P_j| - 1} \mathbb{I}(K_{j,i} < K_{j,i+1}), & |P_j| > 1 \\
0.5, & |P_j| = 1 
\end{cases}
\end{equation}

The overall KSC across all \( N \) paths is calculated by averaging the individual \( KSC_j \) scores, as shown in Equation~\eqref{ksc_ave}. The value of \( KSC \in [0,1] \), where higher scores indicate better alignment with progressive knowledge sequencing.

\begin{equation}\label{ksc_ave}
KSC = \frac{1}{N} \sum_{j=1}^{N} KSC_j
\end{equation}

\subsection{Experimental Strategy}

% All experiments were conducted on a high-performance computing server equipped with a 2.2GHz processor, 64GB of RAM, and dual GPUs (each with 16GB of VRAM). The experimental environment includes Python 3.8, PyTorch 1.13.1, and Ubuntu 18.04.6 as the operating system. 

The experimental workflow consists of the following four components: 

(1) Information extraction for learning path planning.  
We extracted the necessary information required for personalized path generation using existing technical modules. This includes the identification of at-risk students, their knowledge states, preliminary resource lists, video content metadata, knowledge point information, and learner background profiles. These data form the foundation for personalized learning path construction. 

(2) Comparison of foundation model performance. Using the proposed multi-agent planning method, we evaluated the performance of various LLMs in the learning path planning task. This comparison aimed to determine the most suitable foundation model for subsequent experiments. 

(3) Comparison with baseline models. After selecting the best-performing LLM, we performed a comprehensive evaluation using the full proposed method. Its performance was compared against two baseline models (RBM and SLMLPP) to assess the effectiveness of the multi-agent framework. 

(4) Ablation study. To quantify the contribution of each module in the proposed approach, we conducted ablation experiments by removing specific components from the full framework. The tested variants include: removing the learner analytics agent (No\_Analytics), the reflection agent (No\_Reflection), the cognitive load constraint module (No\_CLT), the zone of proximal development constraint module (No\_ZPD), and both constraint modules together (No\_CLT\_ZPD). The resulting performance changes help assess the necessity and impact of each component, and validate the overall effectiveness of the method.

% To promote reproducibility and transparency, all experimental data and source code used in this study have been made publicly available at: \url{https://github.com/xingyezn/MALPP}. Researchers are encouraged to build upon this work for further development and collaborative progress in the field.

\section{Results and Analysis}

\subsection{Information Extraction Results for Learning Path Planning}

\begin{figure}[h]
  \centering
  \includegraphics[width=\columnwidth]{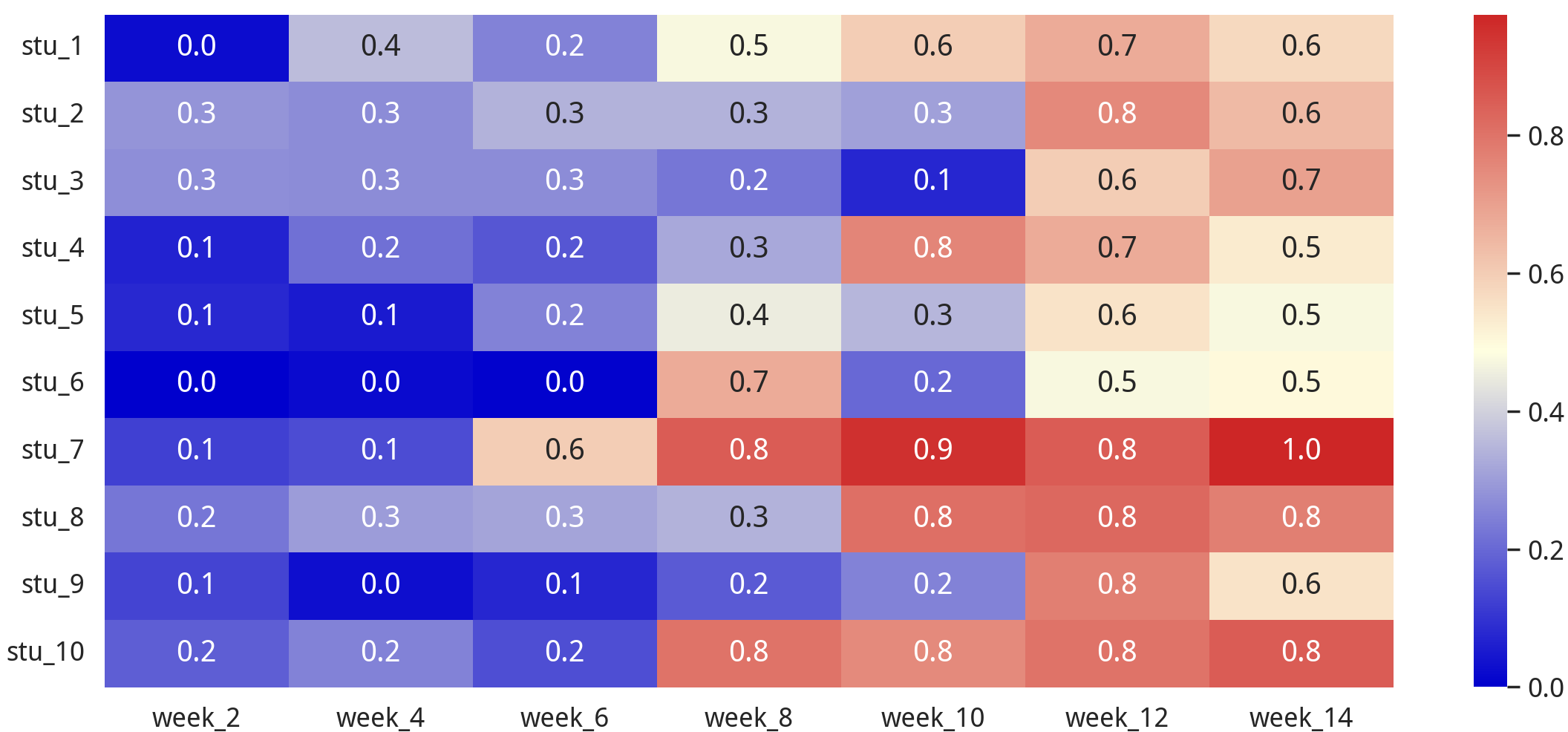}
  \caption{Visualization of early identification results for at-risk students}
  \label{risk_stu}
\end{figure}

We first performed academic risk prediction for all students using the MFFEP model, which outputs the probability of each student being classified as academically at risk for each week. The probability ranges from 0 to 1, with higher values indicating a greater likelihood of academic risk. In this experiment, students with a weekly academic risk probability greater than 0.5 were identified as at-risk, and the corresponding weeks were recorded. Figure~\ref{risk_stu} visualizes the weekly changes in predicted academic risk probability, where blue indicates low risk and red indicates high risk. Students marked in red require timely interventions to reduce the likelihood of academic failure. For example, Student 4 was identified as at risk in Week 10, suggesting that intervention should be implemented at this point. Similarly, Student 7 had a risk probability of 0.6 in Week 6, with the probability continuing to rise in subsequent weeks, indicating that Week 6 would be an optimal intervention window.

\begin{figure}[H]
  \centering
  \includegraphics[width=\columnwidth]{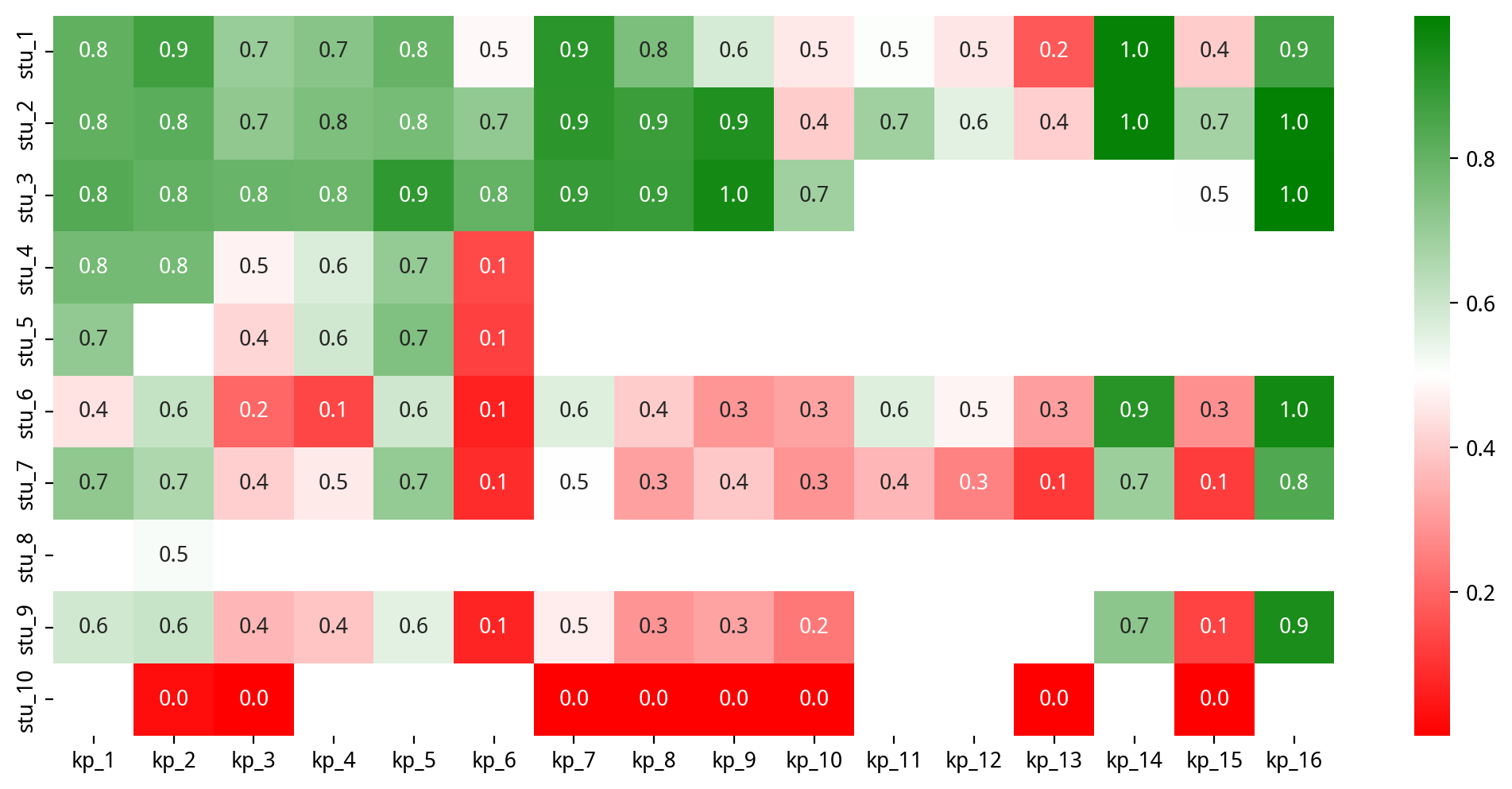}
  \caption{Visualization of knowledge states for at-risk students}
  \label{risk_tu_kp}
\end{figure}

After determining the warning week for at-risk students, we extracted their answering records prior to that week and generated supplementary learning data. Using the SIKT model, we diagnosed the students’ knowledge states at the warning week and calculated their mastery levels for knowledge points related to the enrolled courses, with values ranging from 0 to 1. Figure~\ref{risk_tu_kp} shows the knowledge states of at-risk students in the course “Fundamentals of Mechanical Design,” where red indicates weak knowledge points that require targeted reinforcement, green denotes mastered knowledge points, and gray indicates unlearned knowledge points. Knowledge tracing analysis allows us to precisely identify each student’s mastery gaps, providing a solid basis for personalized learning interventions. For instance, Student 3 had mastered all the knowledge points they had studied, with only some remaining unlearned content; in this case, we recommend assigning new learning resources to accelerate progress. In contrast, Students 6 and 7 had studied all knowledge points in the course but still exhibited substantial knowledge gaps, necessitating personalized learning paths and targeted reinforcement strategies.

Finally, to support learning path planning for at-risk students, we applied the LRRA-LBP model to generate a list of recommended learning resources. For each at-risk student, a preliminary recommendation list of length 20 was produced, serving as the basis for subsequent path planning. At this point, all the key information required for learning path planning has been successfully extracted, enabling the next phase of personalized learning path generation.

\subsection{Evaluation of Different Foundation Models}

To select the most suitable foundation model, we evaluated seven large language models (LLMs), including GPT-3.5, GPT-4o, Llama3.3, Claude-3, DeepSeek-V3, Qwen-Plus, and XXNU-Plus, using the proposed MALPP method. The evaluation results are shown in Table~\ref{diff_llm_RE}, focusing on four core metrics (APL, ALD, KSC, and CLMR) to comprehensively assess each model’s learning path planning capability.

As shown in Table~\ref{diff_llm_RE}, there are notable differences in APL among the models, while ALD varies only slightly. Specifically, in terms of APL, GPT-4o and Qwen-Plus generated the longest learning paths, with mean values of 5.57 and 5.21, respectively. XXNU-Plus, Llama3.3, and DeepSeek-V3 produced moderately long paths (4.78, 4.76, and 4.64, respectively), while Claude-3 and GPT-3.5 generated the shortest paths (3.85 and 2.97, respectively). In contrast, ALD values ranged from 4.40 to 4.96 minutes across all models. XXNU-Plus had the shortest ALD at 4.40 minutes, suggesting more concise recommended resources, whereas Qwen-Plus had the longest ALD at 4.96 minutes, potentially reflecting more detailed learning content but also a relatively higher learning burden. These findings indicate that different foundation models exhibit distinct path planning styles, varying in both path length preferences and average resource duration.

\begin{table}[h] % 改成 h 或 tbp，不用 H
\caption{Learning path planning results of different foundation models}
\centering
\small
\begin{tabular}{lllll}
\toprule
\textbf{Models} & \textbf{APL} & \textbf{ALD} (\textit{mins}) & \textbf{KSC} (\%) & \textbf{CLMR} (\%)\\
\midrule
GPT-3.5 & 2.97 & 4.82 & 54.74 & 36.20 \\
GPT-4o & 5.57 & 4.83 & 61.33 & 39.13 \\
Llama3.3 & 4.76 & 4.90 & 60.45 & 36.24 \\
Claude-3 & 3.85 & 4.58 & 55.33 & 44.23 \\
Qwen-Plus & 5.21 & 4.96 & 58.93 & 45.77 \\
DeepSeek-V3 & 4.64 & 4.74 & \textbf{64.73} & 44.75 \\
XXNU-Plus & 4.78 & 4.40 & 61.94 & \textbf{32.07} \\
\bottomrule
\end{tabular}
\label{diff_llm_RE}
\end{table}

In terms of KSC, DeepSeek-V3 achieved the highest score (64.73\%), demonstrating strong logical and sequential consistency in knowledge structuring. XXNU-Plus and GPT-4o also performed well, with KSC values of 61.94\% and 61.33\%, respectively. By comparison, GPT-3.5, Qwen-Plus, and Claude-3 all scored below 60\%, while Llama3.3 barely exceeded 60\%, indicating room for improvement in maintaining progressive difficulty in knowledge sequencing. Regarding CLMR, XXNU-Plus achieved the lowest value (32.07\%), indicating the highest precision in aligning learning path difficulty with learners’ cognitive load capacities. In contrast, Claude-3, Qwen-Plus, and DeepSeek-V3 all had CLMR values above 44\%, while GPT-3.5, GPT-4o, and Llama3.3 also exceeded 36\%. Higher CLMR values may suggest insufficient consideration of learners’ cognitive load levels, potentially leading to mismatches in task difficulty. Overall, these results reveal that different foundation models excel in different aspects. DeepSeek-V3 performed best in knowledge sequence consistency, making it suitable for learning scenarios that require strong structural coherence. XXNU-Plus achieved the best cognitive load alignment, making it ideal for tasks that demand high sensitivity to learners’ cognitive capacities. GPT-4o, with its tendency to produce longer learning paths, is better suited for scenarios requiring more comprehensive content coverage.

After comprehensive consideration, XXNU-Plus was selected as the foundation model for this study for two main reasons. First, it demonstrated the best CLMR performance and the second-best KSC score (only behind DeepSeek-V3), offering a balanced capability profile. This may be attributed to XXNU-Plus being an optimized integration of multiple models, achieving competitive overall performance. Second, the multi-agent method used in this study consumes a substantial number of tokens (approximately 20,000 per generated path). From a computational cost perspective, XXNU-Plus, being locally deployed, offers a more cost-effective solution compared to cloud-based models requiring paid API access, making it a practical choice for balancing performance and resource efficiency.

\subsection{Performance Comparison of Different Baseline Models}

After selecting XXNU-Plus as the foundation model, we compared the performance of different baseline approaches. The results are shown in Table~\ref{path_RE}. It can be observed that the RBM method achieved higher APL and ALD values than the other two methods. This is because its path lengths were randomly generated between 1 and 10, resulting in an APL of 5.10, while its ALD was 5.58 minutes, which is close to the average learning duration across all resources. However, RBM achieved only 48.35\% in KSC and 53.81\% in CLMR, both lower than those of SLMLPP. Moreover, RBM cannot provide explanations for its recommendations, making it unsuitable for highly personalized learning needs.

In contrast, the SLMLPP method, although lacking a multi-agent architecture, can provide explainable path recommendations. Its KSC score improved by 2.64\% compared to RBM, but was still 10.95\% lower than that of MALPP. Furthermore, SLMLPP performed worst in CLMR, with a misalignment rate of 57.61\%, significantly higher than MALPP’s 32.07\%. This indicates that SLMLPP failed to effectively match the learning path to the learner’s cognitive load, resulting in a higher cognitive load misalignment rate. These findings suggest that relying solely on a single LLM for path planning can produce explainable paths but may lead to limitations in cognitive load adaptation and knowledge sequence consistency. In contrast, the proposed MALPP method incorporates Cognitive Load Theory (CLT) and Zone of Proximal Development (ZPD) constraints into the agent design and leverages multi-agent collaboration, ensuring that the generated paths are both reasonable and adaptable.

\begin{table}[h]
\caption{Learning path planning results of different baseline models}
\centering
\small
\begin{tabular}{lllll}
\toprule
\textbf{Models} & \textbf{APL} & \textbf{ALD} (\textit{mins}) & \textbf{KSC} ($\%$) & \textbf{CLMR} ($\%$)\\
\midrule
RBM & 5.10 & 5.58 & 48.35 & 53.81 \\
SLMLPP & 4.82 & 5.46 & 50.99 & 57.61 \\
MALPP & 4.78 & 4.40 & \textbf{61.94} & \textbf{32.07}\\
\bottomrule
\end{tabular}
\label{path_RE}
\end{table}

\begin{figure}[H]
  \centering
  \includegraphics[width=0.9\columnwidth]{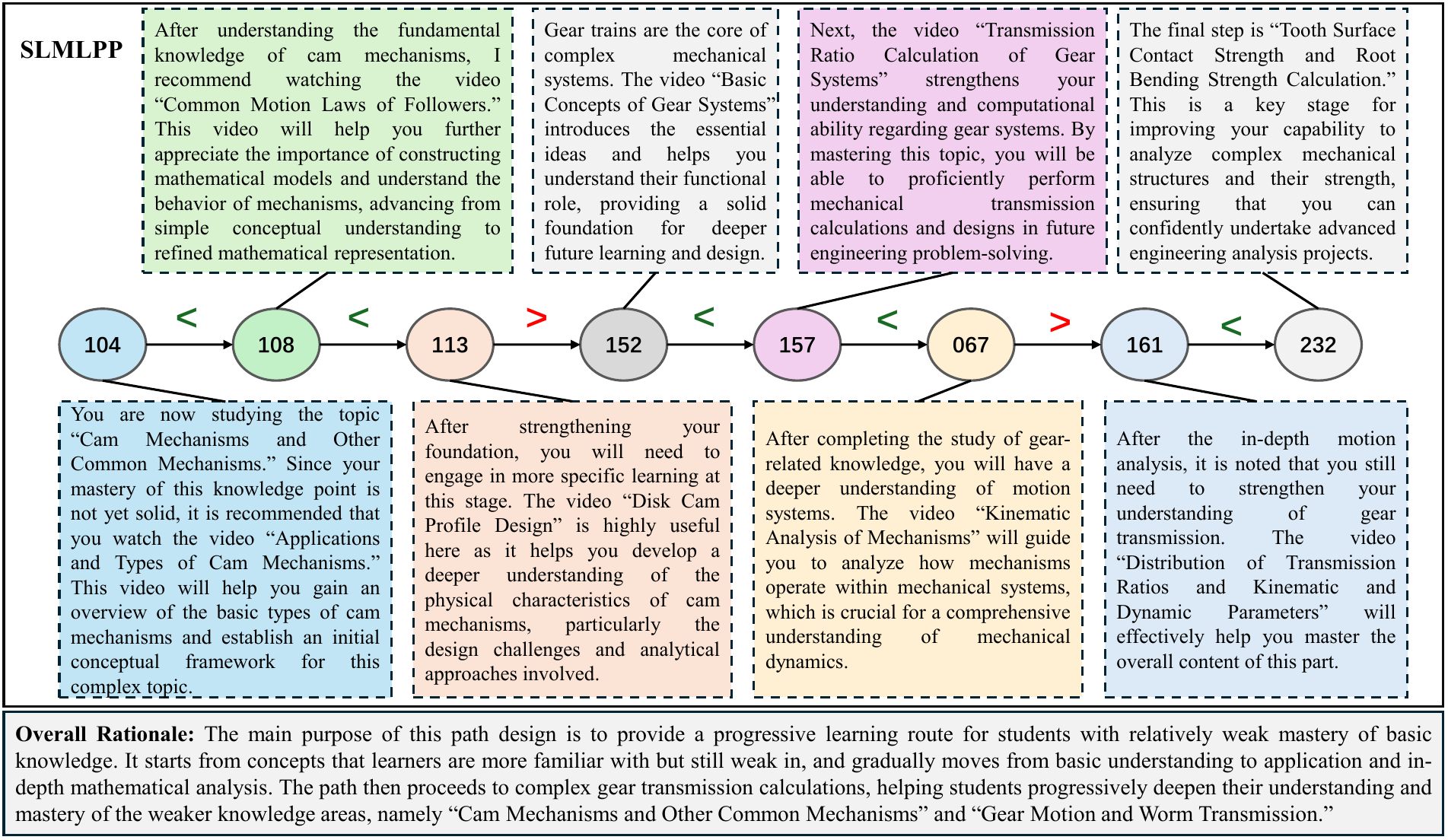}
  \caption{Example learning path generated by SLMLPP}
  \label{SLMLPP}
\end{figure}

\begin{figure}[H]
  \centering
  \includegraphics[width=0.9\columnwidth]{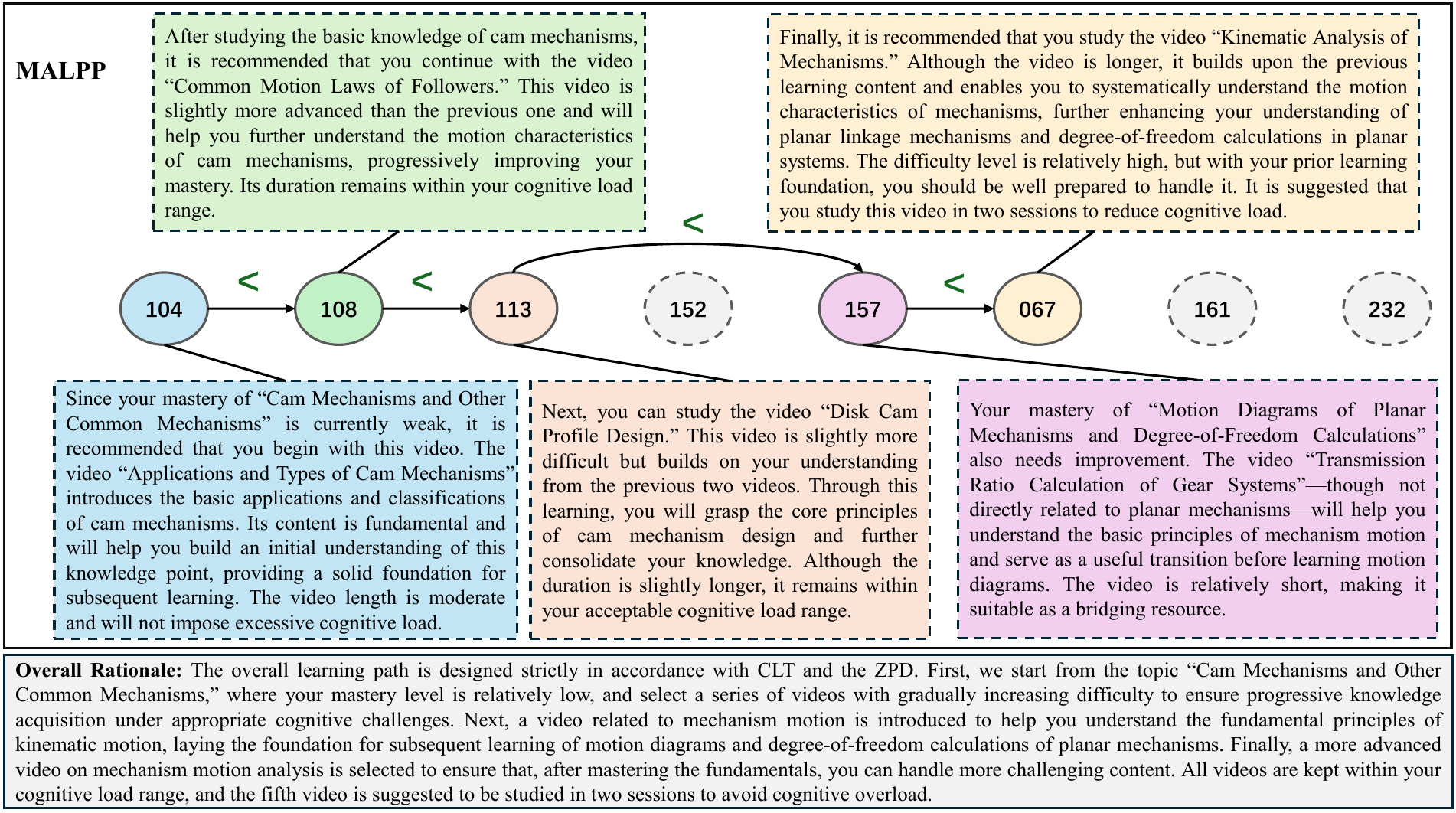}
  \caption{Example learning path generated by MALPP}
  \label{MALPP}
\end{figure}

To further illustrate the differences between SLMLPP and MALPP, we conducted a case study using a student enrolled in the “Fundamentals of Mechanical Design” course who was identified as at risk in Week~8. Figures~\ref{SLMLPP} and~\ref{MALPP} show the learning paths generated by SLMLPP and MALPP, respectively, for this student. In each diagram, ellipses represent different learning resources, dashed rectangles above and below indicate recommendation rationales, and the greater-than/less-than symbols between resources indicate difficulty comparisons.

As shown in Figure~\ref{SLMLPP}, the SLMLPP method recommended a path length of 8, close to the maximum allowed length of 10, which may exceed the student’s cognitive load capacity. Moreover, the recommended resources were not strictly organized in a progressive manner, and the rationales did not sufficiently consider cognitive load factors, potentially leading to imbalanced path difficulty. In contrast, as shown in Figure~\ref{MALPP}, the MALPP method produced a path length of 5, skipping relatively easy resources 152 and slightly more difficult resources 161 and 232, while recommending that resource 067 be studied in two separate sessions. This ensured that the recommended resources did not exceed the learner’s cognitive capacity. Furthermore, the resources in the MALPP path were arranged in ascending difficulty, enabling the learner to acquire knowledge progressively, thus improving both efficiency and adaptability.

In summary, while SLMLPP can produce explainable learning paths, it shows limitations in cognitive load adaptation and knowledge sequence consistency, which may lead to higher cognitive load misalignment rates. By incorporating CLT and ZPD constraints and leveraging a multi-agent collaboration mechanism, MALPP generates learning paths that are more reasonable, better adapted to the learner, and more effective at reducing cognitive load misalignment, thereby improving learning efficiency. This demonstrates the significant value of introducing multi-agent collaboration into learning path planning for enhancing personalized learning outcomes.

\subsection{Ablation Study Results}

To verify the contribution of each component in the MALPP method, we conducted an ablation study, and the results are presented in Table~\ref{abla_re_path}.

First, regarding the APL and ALD metrics, all methods showed relatively similar APL values, likely because they share the same foundation model, resulting in minor differences in path length selection. However, in terms of ALD, the No\_CLT and No\_CLT\_ZPD methods recorded noticeably higher values than the other methods. This is because removing the cognitive load constraint module caused the model to fail to distinguish learning durations when selecting resources, thereby reducing the rationality of the generated learning paths. Second, in terms of KSC and CLMR, the No\_Analytics method, which removes the learner analytics agent, did not show significant performance degradation compared to the original MALPP in either metric. This may be because the path planning agent also performs a degree of learner analysis during its execution, partially compensating for the removal of the analytics agent. Nevertheless, we consider this module valuable, as it produces detailed learner analysis reports for both students and instructors, helping them better understand learning progress.

\begin{table}[h]
\caption{Ablation results of the MALPP method}
\centering
\small
\begin{tabular}{lllll}
\toprule
 & \textbf{APL} & \textbf{ALD} (\textit{mins}) & \textbf{KSC} ($\%$) & \textbf{CLMR} ($\%$)\\
\midrule
No\_Analytics & 4.56 & 4.80 & 61.07 (-0.87) & 32.56 (-0.49)\\
No\_Reflection & 4.68 & 4.90 & 61.11 (-0.83) & 39.92 (-7.85)\\
No\_CLT & 4.84 & 5.34 & \textbf{62.86 (+0.93)} & 45.32 (-13.25)\\
No\_ZPD & 4.03 & 4.65 & 56.74 (-5.19) & 43.03 (-10.96)\\
No\_CLT\_ZPD & 4.97 & 5.80 & 55.54 (-6.40) & 53.31 (-21.24)\\
\midrule
MALPP & 4.78 & 4.40 & 61.94 & \textbf{32.07}\\
\bottomrule
\end{tabular}
\label{abla_re_path}
\end{table}

For the No\_Reflection method, removing the reflection agent resulted in learning paths being generated only once without further optimization, leading to a 7.85\% decrease in CLMR performance and a 0.83\% decrease in KSC. This suggests that the reflection agent plays a key role in enhancing the rationality of the learning paths. Further analysis of the removal of the CLT and ZPD constraint modules shows that the No\_CLT\_ZPD method experienced a 6.40\% drop in KSC and a substantial 21.24\% drop in CLMR, highlighting the critical role of these two modules in improving both knowledge sequence consistency and cognitive load alignment. Removing only the CLT module (No\_CLT) caused the model to rely more heavily on ZPD constraints, leading to a 13.25\% drop in CLMR but a slight 0.93\% increase in KSC, indicating that the CLT module primarily contributes to improving overall path adaptability. On the other hand, removing only the ZPD module (No\_ZPD) caused a 5.19\% decrease in KSC, which was larger than in the No\_CLT case, and a 10.96\% drop in CLMR, which was smaller than that of No\_CLT. This further confirms the ZPD module’s importance in optimizing the knowledge sequence.

In summary, the ablation experiments validate the effectiveness of all components in the MALPP method. In particular, the CLT and ZPD constraints are essential for enhancing both knowledge sequence consistency and cognitive load alignment, while the reflection agent significantly improves the overall rationality of the generated learning paths.

\section{Discussion}\label{sec12}

The experimental results demonstrate that the proposed MALPP method effectively improves both the adaptability and explainability of learning path planning compared to baseline approaches. The integration of CLT and ZPD constraints plays a pivotal role in aligning learning paths with learners’ cognitive capacities and ensuring progressive knowledge sequencing. In particular, the ablation study confirms that removing either of these constraints results in substantial performance degradation in CLMR and KSC, highlighting their necessity for achieving high-quality path recommendations.

Compared with existing single-agent LLM-based approaches, MALPP leverages a distributed multi-agent collaboration framework that decomposes the path planning process into specialized subtasks, including learner analytics, path generation, and reflective evaluation. This design not only enhances the interpretability of the generated paths but also allows for iterative optimization through agent interaction. The case study further illustrates that MALPP can produce more balanced and cognitively appropriate paths, avoiding excessive length and difficulty jumps that might hinder learner engagement.

From a theoretical perspective, the incorporation of CLT and ZPD into the prompt engineering process represents a novel approach to embedding pedagogical principles into LLM-driven educational systems. Practically, MALPP offers an adaptable framework that can be extended to various learning contexts, such as different subject domains, online course formats, or adaptive assessment systems. Nevertheless, some limitations should be acknowledged. First, the method relies on high-quality learner interaction data for accurate profiling and recommendation, which may limit its applicability in scenarios with sparse data. Second, the computational cost of multi-agent interactions is non-trivial, although local deployment of the foundation model can partially mitigate this issue. Future research could explore model compression techniques or hybrid architectures to further reduce resource consumption. Additionally, integrating multimodal resources into the planning process could improve recommendation richness and engagement.

\section{Conclusion}\label{sec13}

This study proposed a multi-agent driven, explainable learning path planning method (MALPP) that integrates large language models with pedagogical constraints derived from Cognitive Load Theory and the Zone of Proximal Development. The method combines learner analytics, knowledge state tracking, personalized resource recommendation, and iterative path reflection to produce adaptive and interpretable learning paths. Extensive experiments on the MOOCCubeX dataset demonstrated that MALPP outperforms baseline methods in both knowledge sequence consistency (KSC) and cognitive load misalignment rate (CLMR), while maintaining competitive performance in average path length (APL) and average learning duration (ALD). The ablation study further confirmed the critical role of the CLT and ZPD constraint modules, as well as the reflection agent, in ensuring the rationality and adaptability of generated learning paths. The proposed approach offers both theoretical and practical contributions: it embeds well-established educational theories into LLM-based path planning, and it provides a modular architecture that can be extended to various adaptive learning systems. Future work will focus on incorporating richer learner models, supporting multimodal learning resources, and exploring real-time path adaptation to further enhance the personalization and effectiveness of intelligent tutoring systems.

\backmatter

\bibliography{sn-bibliography}% common bib file

\end{document}